\documentclass[runningheads]{llncs}
\usepackage[T1]{fontenc}
\usepackage{graphicx}
\usepackage{booktabs}
\usepackage[misc]{ifsym}
\newcommand{\corr}{(\Letter)}

\usepackage[utf8]{inputenc}
\usepackage{amsmath, amssymb}
\usepackage{xcolor}
\usepackage{hyperref}
\hypersetup{colorlinks, linkcolor=black, citecolor=black, urlcolor=blue}
\usepackage{microtype}
\usepackage{caption}

\usepackage[textsize=small,backgroundcolor=green]{todonotes}

\definecolor{MyraFGcolor}{RGB}{0,130,50} 
\definecolor{onemoreFGclor}{RGB}{00,150,00}
\definecolor{thirdFGcolor}{RGB}{00,150,80}

\definecolor{finFGcolor}{RGB}{0,104,180} 
\definecolor{SzymonFGcolor}{RGB}{255,00,170}
\definecolor{HamzaFGcolor}{RGB}{20,50,190}
\definecolor{blue3FGcolor}{RGB}{00,100,220}

\definecolor{todoinBG}{RGB}{255,130,0} 
\definecolor{paleBG}{RGB}{220,220,220}
\definecolor{kmdBGcolor}{RGB}{152,251,152} 
\definecolor{otherTeamBGcolor}{RGB}{175,238,238} 

\begin{document}

\title{Counterfactual Transition Graphs:\\
Evaluating Cross-Class Transition Quality
  }
\titlerunning{Counterfactual Transition Graphs}

\author{Syed Muhammad Hamza Zaidi\inst{1} \corr \and
Szymon Bobek \inst{2} \and
Grzegorz J. Nalepa\inst{2,3} \and
Myra Spiliopoulou \inst{1} 
}
\authorrunning{S. M. H. Zaidi et al.}
\institute{Otto-von-Guericke University Magdeburg, Magdeburg, Germany \email{syed.zaidi@ovgu.de}
\and
Jagiellonian University, Krakow, Poland
\and Center for Applied Intelligence Systems Research, School of ITE, Halmstad University, Halmstad 30118, Sweden
}

\maketitle

\begin{abstract}

Counterfactual (CF) explanations for time-series classifiers are usually
evaluated one example at a time:
\emph{what minimal edit flips this single window's prediction?}
We argue that the more informative question for diagnostic
interpretability is structural: \emph{how does the classifier connect
its own classes to each other?}

We propose a \textbf{counterfactual transition graph} (CGT) in which each node
is a class and each edge weight is the CF reliability of the transition
from one prototype to another under a proximity aware retrieval sweep.
On a six-class hand-movement task, we induce a CGT that
reveals a non-trivial topology, which
is \emph{not} predicted by the binary confusion matrix: it shows that
counterfactual reachability does not align with classifier accuracy and
even \textit{runs counter to it} (Spearman $\rho=-0.37$ over the 15 pairs),
i.e. the boundaries the classifier separates most confidently are among
those an in-distribution edit can least often cross.

Our framework is method agnostic, i.e. any CF-explainers can be used. Presently, we use it to juxtapose replacement-based CFs with gradient-based CFs; gradient-based methods reach almost any class by
stepping off the data manifold, while replacement-based methods stay
on it and fail on precisely the rigid boundaries.

\end{abstract}

\section{Introduction}\label{sec:intro}

A counterfactual (CF) explanation asks a \emph{local} question: what is the smallest change to an input that flips a classifier's prediction to a chosen class? ~\cite{guidotti2024cf}. The edit should keep the instance plausible, so a bird flipped from `crested auklet' to
`red-faced cormorant'~\cite{GoyalEtAl:ICML2019} still looks like a bird, and a hand-movement window edited toward another gesture still looks like real motion. Instead of studying one instance and its CF at a time, we
ask a \emph{global} question:
what does the whole set of class-to-class edits reveal about the (im)plausibility of counterfactuals?

\begin{figure}[!htb]
  \centering
  \includegraphics[width=0.92\linewidth]{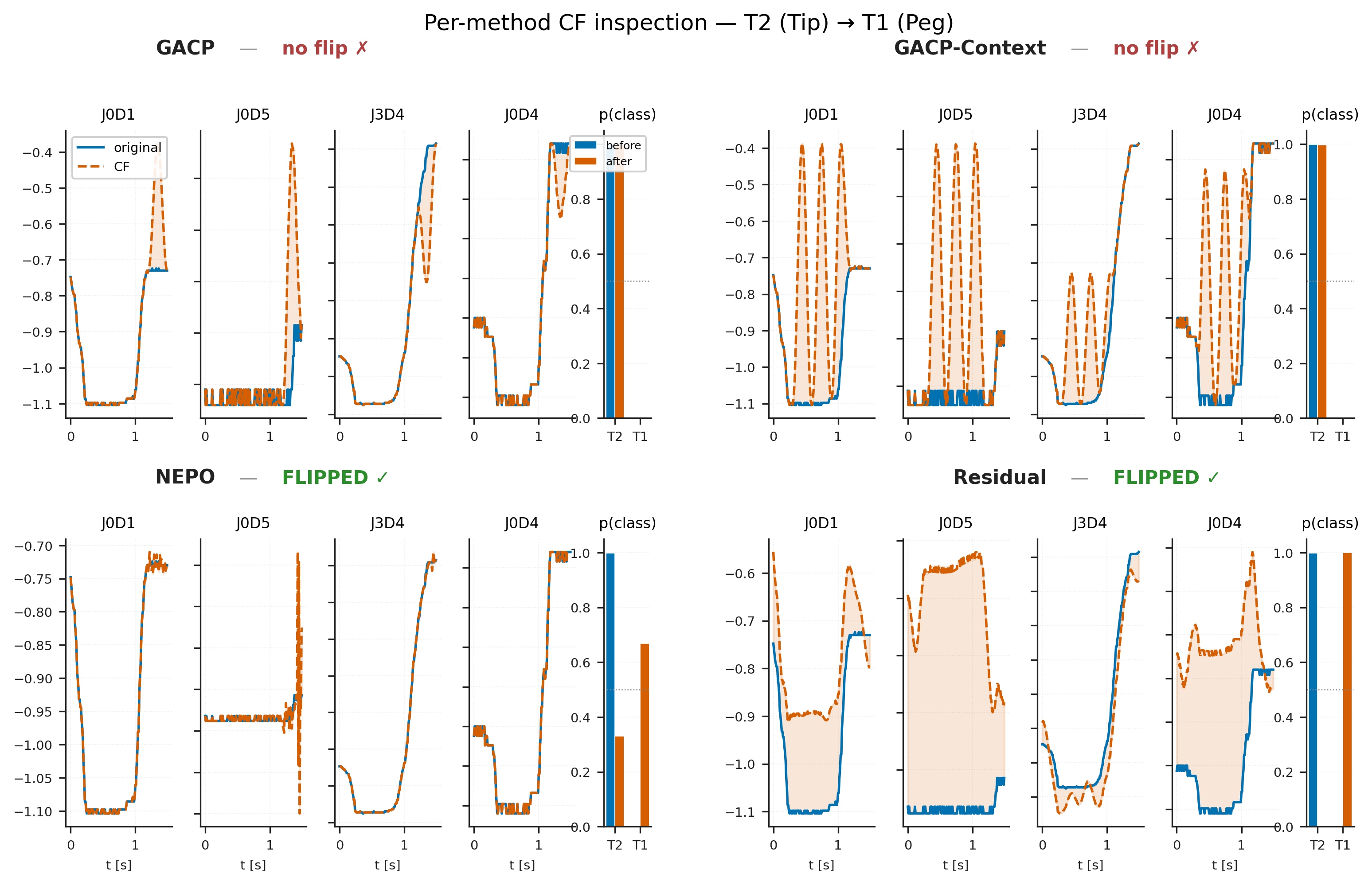}
  \caption{Transitions of a hand's joints from the class 
  Fingertip-Touch ($T_2$) to the class Clothes-Peg ($T_1$):
  each block shows four representative joints (original: blue, counterfactual: orange) and the class probabilities before and after the edit. The replacement-based generators (\texttt{gacp}, \texttt{gacp\_context}) do not flip the prediction, while
  \texttt{gacp\_context} produces physiologically implausible oscillations;
  the gradient-based generators (\texttt{nepo}, \texttt{residual}) flip the
  label with a small but
  off-manifold edit. 
  }
  \label{fig:badcf}
\end{figure}

Figure~\ref{fig:badcf} shows a challenging case from our data: each subfigure shows the edits on the finger joints of a hand to reach the counterfactual (orange) from the original instance (blue).
We explain Figure \ref{fig:badcf} later. For now, we would like to point to the two rightmost subfigures:
while the lower one depicts smooth movements, the upper one shows jumps that no human finger joint can perform.

\noindent
Inspired by this example, we formulate following research questions:

\begin{description}
\item [RQ1:] How can we capture the (im)plausibility of a counterfactual?
\item [RQ2:] How can we link it to the (dis)similarity between the class of the
  instance and the class of the counterfactual? 
\item [RQ3:] How can we validate counterfactual (im)plausibility in a
  controlled way?
\end{description}

For RQ1 and RQ2 we represent each class by  a number of 
\emph{prototypes}, defined as
recordings chosen as the most typical instances of that class (Section~\ref{sec:method}), 
and we record for every class pair how reliably a prototype-anchored counterfactual carries one class to the other, aggregated over folds, prototype pairs and retrieval strategies.
The resulting \emph{Counterfactual Transition Graph} (CFTG) has one edge weight per pair.
For RQ3, we work with predefined classes and study how the (dis)similarity among the prototypes is reflected in counterfactual quality.
We report on our results for
a finger gestures dataset, where we found that the most strange counterfactuals are between classes separated with high confidence.


Our contributions are as follows.
(1)~We define the counterfactual transition graph, a class-level structure built
by aggregating prototype-to-prototype counterfactuals under a prototype-anchored
retrieval scheme that makes the aggregation well-posed (Section~\ref{sec:method}).
(2)~On a temporal graph classifier 
we show that counterfactual reachability does not track accuracy, and if anything the two run counter: the instances the classifier separates most confidently are among the hardest to reach
by an in-distribution counterfactual (Section~\ref{sec:results}).
(3)~Because the construction is generator-agnostic, the same graph exposes
\emph{method dependence}: gradient-based and replacement-based generators disagree
exactly where one of them edits the data manifold, which separates conclusions
about the classifier from artefacts of the explainer (Section~\ref{sec:results}).

This study is grounded in the SenseGlove hand-movement use case, which we treat as a challenging reference setting. It combines multimodal joint readings, temporal dependencies, and several movement classes with different kinematic characteristics, making it suitable for studying whether counterfactual transitions reveal structure beyond single-instance explanations and classification accuracy. We therefore propose the CFTG as a general diagnostic idea, but evaluate it here as a solution motivated by this concrete use case; testing its transfer to other domains remains future work.

Our paper is designed as follows. In Section~\ref{sec:related} we discuss related work.
In Section~\ref{sec:materials} we present our materials: the SenseGlove hand-movement dataset, our temporal graph classifier for it, and the four counterfactual generators we use.
Section~\ref{sec:method} contains our approach: the prototypical class representation, the notion of class proximity, the per-pair transition operation, and their aggregation into the CFTG. In Section~\ref{sec:setup} we describe the experimental setup and in Section~\ref{sec:results} our results. Section~\ref{sec:disc} closes the paper with summary and outlook.

\section{Related Work}\label{sec:related}
Related to our work are counterfactual-based explanations for time series mechanisms for evaluating them.

\subsection{Counterfactual-based explanations in time series}
There is broad literature on explaining time series with counterfactuals, and recent reviews highlight recurring design patterns \cite{schlegel2026whatif}. We organize this research into three groups, 
primarily by the \emph{source of the edit}, that is, by whether the counterfactual is generated by directly perturbing the queried instance, by substituting parts of it with patterns taken from other instances, or by retrieving a nearby unlike example that already serves as a counterfactual candidate.

The first group comprises \emph{perturbation-based} methods, in which the queried time series is the main source of the counterfactual, and the explanation is generated by directly modifying this instance until the model prediction flips. A representative example is Glacier, which performs guided and locally constrained optimization in the input space to obtain counterfactuals that remain close to the original signal while changing the classifier's decision~\cite{Wang2024GlacierGLM}. Methods such as M-CELS follow the same general principle, but use a learned saliency map to focus the perturbation on temporally informative regions, thereby improving sparsity and interpretability of the edits~\cite{Li2024MCELSCEH}.

The second group comprises \emph{substitution-based} methods, in which the counterfactual is still obtained by modifying the original time series, but the modification is performed through the insertion, replacement, or patching of subsequences derived from other examples. A representative method in this family is motif-guided counterfactual generation, where discriminative subsequences of the query are replaced with motifs drawn from instances of the target class~\cite{Li2023MotifguidedTSK}. Shapelet-based methods follow a closely related idea, but use discriminative shapelets to
guide the replacement process, making the transition from one class to another more structured; these methods still rely on external temporal patterns rather than free-form perturbation of the original signal~\cite{Bahri2022ShapeletBasedCEL}. Prototype- or symbolic-pattern-based approaches also
construct the explanation through representative temporal fragments or concepts originating from other instances~\cite{Mozolewski2026FromPTG,Pludowski2025MASCOTSMSD}.

The third group comprises \emph{native} methods, where the counterfactual is itself a nearest unlike neighbour, or a slightly adapted version of such a neighbour~\cite{delaney2021native}, rather than being synthesised from scratch.

\subsection{Explainers in kinetics}

We use the kinetics dataset as use case, hence dedicated
studies on CF for kinetics are not directly relevant. But one insight is, namely that the labels themselves may be uncertain:
Dindorf et al point out that 
for the dataset and two diagnostic tasks they studied (hyperlordosis and hyperkyphosis), 'around 10\% of the test labels were incorrect' \cite{DindorfEtAl:Bioengineering2023}. Since we want to evaluate the CFs themselves, we do not choose diagnostic tasks in downstream classification, but rather the gestures themselves as classes.

\color{black}

\subsection{How to evaluate counterfactual explainers?}

Counterfactual explainers are typically evaluated along several key dimensions~\cite{theissler2022survey,guidotti2024cf}, most notably validity (i.e., whether the generated instance indeed changes the model's prediction), proximity (the magnitude of the change between the original instance $x$ and the counterfactual $x'$), and plausibility, often operationalized via data manifold constraints, density estimation, or reconstruction error. 

For time-series data, these evaluation dimensions are extended with constraints specific to temporal signals, such as smoothness, continuity, temporal  coherence and preservation of signal dynamics. To this end, domain-aware metrics -- e.g., dynamic time warping (DTW), spectral similarity, or shape-based distances -- are commonly employed to assess the realism of generated sequences. Despite these advances, current evaluation approaches largely focus on instance-level properties and often do not explicitly account for differences between pairs of classes, ie differences that are expected to appear on each member (and built-up counterfactual) of a class.

An orthogonal concept is
actionability~\cite{karimi2022recourse}, which refers to modifications that can be implemented in a real-world setting, for example by avoiding changes to immutable features and preserving semantic or logical relationships between features. 
Although some works consider the plausibility of the transition itself, they typically aim to provide general frameworks for assessing this property rather than analyzing the pairwise difficulty of transitions between specific classes~\cite{Poyiadzi2019FACEFA}.


\section{Materials}
\label{sec:materials}
For our analysis, we use the SenseGlove DK1.3 data-glove
recordings~\cite{sensegloveDK13,Reichert2022SensorsHand,Klemm2024JAPFinger}. Volunteers were wearing this glove and performed pre-specified movements. The ultimate goal is the early recognition of neurodegenerative diseases like Parkinson's through hand movement. The current dataset has only healthy volunteers.

\subsection{Hand activities in the SenseGlove dataset}
\label{sec:senseglove}
The data collection contains
one file per subject ($n=34$, 17 younger and 17 older participants, but we ignore the age group). The file contains subject's trials and their movement-type labels.
These labels are everyday activities: (1)
\emph{Clothes-Peg}, opening and repositioning a spring peg; (2)
\emph{Fingertip-Touch}, the thumb touching each fingertip in turn; and four
cube rotations, a Rubik's cube turned clockwise (3) and counter-clockwise (4),
and a smaller cube turned the same two ways (5, 6). The four rotations are
organised along two intuitive axes, the manipulated object (large or small
cube) and the rotation direction, giving the label set a natural ``size plus
direction'' taxonomy; whether the classifier follows these axes is one of the
questions the transition graph answers in Section~\ref{sec:results}. 

All activities are performed
with the right hand, whose five digits D1 (thumb) to D5 carry four joints
each, giving $N{=}20$ finger-joint angle channels sampled at $100$\,Hz and
indexed as $J\{0,1,2,3\}D\{1,\dots,5\}$. 
Each
participant repeats every activity many times, yielding about $1{,}700$ trials
per activity and roughly $300$ per participant ($\approx\!10{,}180$ in total). For each
trial we derive angular velocity and acceleration and concatenate them
($F{=}3$ features per joint per step), segment the trial into windows of
length $W{=}150$ with stride $75$ ($1.5$\,s at $100$\,Hz), and standardise
features per fold using the training windows only.

\subsection{Encoding a hand's anatomy for movement classification}
\label{sec:pdtg}
\begin{figure}[htb]
  \centering
  \includegraphics[width=\linewidth]{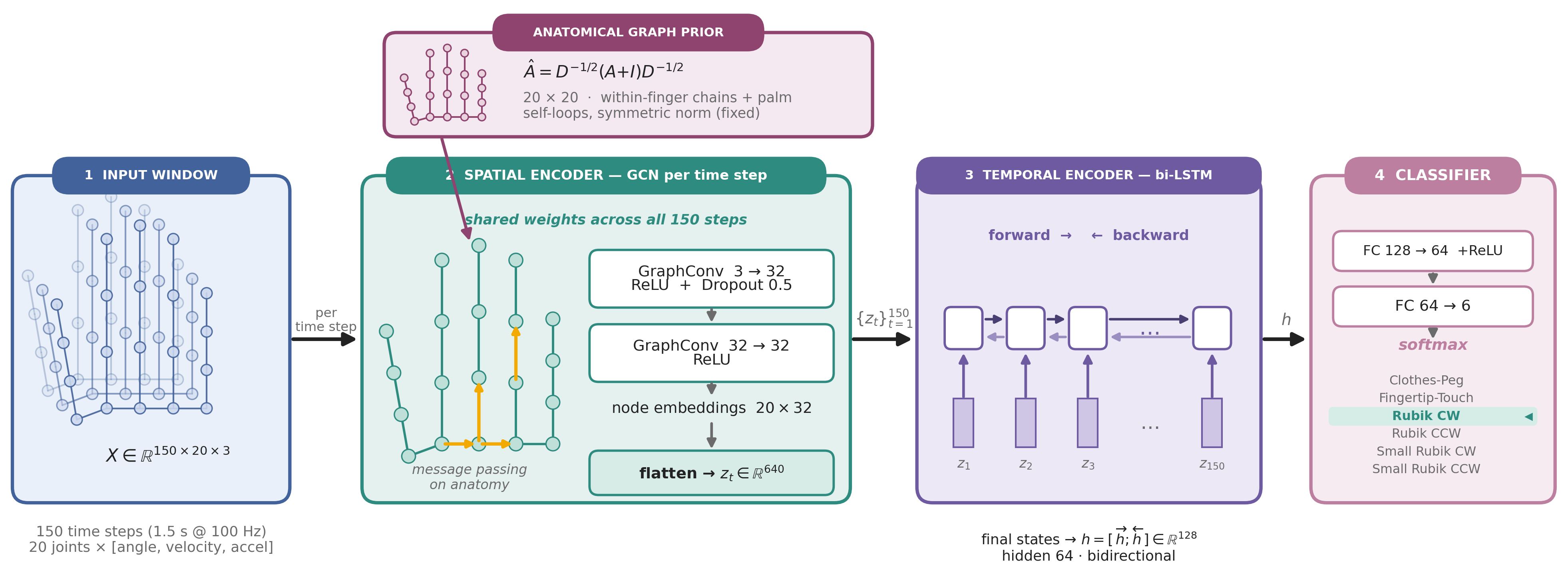}
  \caption{Workflow of the TGNN classifier, summarised
  from~\cite{ZaidiEtAl:CBMS2026}: the temporal sequence of anatomical hands per gesture (subfig 1) are encoded spatially with graph convolutions, each step separately (subfig 2); biLSTM is used for temporal encoding (subfig 3); the MLP-head does downstream classification
}
  \label{fig:workflow}
\end{figure}

To encode the hand's anatomy we represent each window as a signal on a
fixed, undirected anatomical graph, we proposed the Physio-Digital Temporal Graph
(PDTG) \cite{ZaidiEtAl:CBMS2026}: the $20$ joint channels are nodes, the four joints of each digit form a chain, and the base joints are connected across the palm, acting as a built-in prior in the spirit of skeleton graphs for articulated motion ~\cite{Yan2018STGCN}. As base model we used the GNN-LSTM as presented in \cite{ZaidiEtAl:CBMS2026}, and summarized in Figure \ref{fig:workflow}.

As shown in Figure \ref{fig:workflow}, we augment the windowed signals with angles, velocity, and acceleration, and we place them in the anatomical hand graph. We encode each time step spatially with graph convolutions~\cite{kipf2017gcn}, and capture the temporal dimension with a bidirectional LSTM.
We use an MLP head for classification; we train with cross-entropy under subject-grouped cross-validation. Integrated-gradient attributions \cite{Sundararajan2017IntegratedGradients} quantify the contribution of each joint to classification.

\subsection{The counterfactual generation cores}

To produce class transitions under different edit philosophies, we consider 
the methods
direct perturbation (NEPO), local substitution (GACP), context-aware substitution (GACP-Context), and residual attenuation (Residual), as described hereafter. All of them decompose the signal into non-overlapping windows, so the counterfactual transition is done at the window level~\cite{theissler2022survey}. 



\paragraph{Normal-Embedding Prior Optimization (NEPO)}
starts from the original instance and directly optimizes an edited version of the selected windows so that the model prediction moves toward the target class.  The edited regions are chosen using attribution information. The optimization objective combines a classification term with regularization terms encouraging small, sparse, and temporally smooth edits. 
Thus, NEPO does not borrow content from other examples; it rather scans the input space for the smallest perturbation that is sufficient to induce a valid and temporally coherent class transition.

\paragraph{Graph-Aware Contextual Patching (GACP)} is a substitution-based mechanism: 
For each selected window, it searches for a \emph{nearest unlike neighbour}, i.e., a window with similar local structure but belonging to the opposite or desired class, and uses this retrieved segment as the basis for replacement. 
The replacement is then blended into the original signal with smoothing and overlap-aware reconstruction so that the transition remains temporally coherent using graph as a constraint for unrealistic changes. Conceptually, GACP therefore realizes the class change through \emph{window substitution}.

\paragraph{GACP-Context} refines GACP: instead of
retrieving a replacement window only from the similarity of the window itself, it builds a representation of the surrounding neighbourhood and searches for a target-class window whose \emph{context} matches the original one.

\paragraph{Residual:} 
instead of searching for a perturbation or a replacement window, it first decomposes the signal into a learned \emph{shared} component and a residual component interpreted as the deviation that carries class-specific evidence. Counterfactual generation is then performed by \emph{attenuating the residual}: the method progressively suppresses the part of the signal that is responsible for the residual deviation, while preserving the shared morphology of the original instance. 


\section{Our approach}
\label{sec:method}
We turn a per-instance counterfactual (CF) generator into a class-level diagnostic. For each class we build a \emph{prototypical representation} from the windows most characteristic of it, take these prototypes as the nodes of a graph, and weight each edge by how reliably a counterfactual transports one class's prototypes across the decision boundary into another. The resulting weighted graph over the classes is the \emph{Counterfactual Transition Graph} (CFTG). On this graph, we define CF-reachability and CF-rigidity, the concepts we use as basis to assess the interplay between classifier confidence and plausibility of the CFs.




The construction  of CFTG rests on following ingredients: a prototypical class representation (cf. Section~\ref{sec:method_prototypes}), the new notions of class proximity and of transitions between classes (cf. Section~\ref{sec:method_proximity} for both), 
and their aggregation into the CFTG together with the CF-rigidity it exposes (Section~\ref{sec:method_cftg}). The implementation of these ingredients for our evaluation is described in Section~\ref{sec:setup}.



\subsection{Prototypical representation of a class}
\label{sec:method_prototypes}

In explainable AI, a prototype serves as a representative \emph{example} for a given class~\cite{bobek2025tsproto}. For a class $A$ and a fold, we embed each of that class's training windows through the encoder of $f_{AB}$, average its sub-window embeddings into one vector, sort the vectors on proximity to the class centroid (the mean embedding of the class) and pick the top-$m$ proximal vectors. These constitute the set of prototypes $P_A=\{p_{i,A}|i=1\ldots{}m\}$ for class $A$. Conceptually, they are 
\emph{medoid}-like instances of a class in the classifier's representation. They are chosen by typicality rather than by classifier confidence; we implement 'typicality' as proximity to the class-mean embedding, and use cosine similarity as proximity function.

To prevent that a single window dominates, we have $m>1$ medoid-like prototypes per class. To keep the prototypical representation of a class compact, we set $m$ to a small number. In our experiments, we set $m{=}3$.


\subsection{Proximity between classes}
\label{sec:method_proximity}
We distinguish between two concepts of proximity -- the conventional proximity between counterfactuals, and a new kind of proximity at class level. \emph{Counterfactual proximity} refers to the magnitude of the edit required to transform a factual window into its counterfactual; we denote it as 'CF-proximity'. We introduce the concept of \emph{Class-Prototype proximity (CP-proximity)}, intended to quantify the overall structural distance between two classes by using the prototypes in the representation learned by the classifier. We formalize this new form of proximity hereafter, starting with the concept of asymmetric 'transition attempt' from a source class to a target class.

\subsubsection{Transition from one class to another.} 
Let $A$ and $B$ be two classes, let $f_{AB}$ be a classifier trained to separate between them, and let $P_A$ and $P_B$ be their sets of prototypes in the representation space of $f_{AB}$ \footnote{$P_A,P_B$ depend on $f_{AB}$, but we simplify this and write $P_A$ instead of $P^{f_{AB}}_A$.}. Assume that $A$ is the source class and $B$ the target class. Further, let $GC$ be a generator core and $s_{GC}$ be a concrete scan over the solution space of $GC$. The nature of $s_{GC}$ depends on how the generation core searches for solutions and on its optimization criterion. For simplicity, we set $s_{GC}\equiv{}s$ whenever the generation core is fixed. We define:

\begin{itemize}
\item 
a \textit{transition attempt} as a tuple $<(A,B),p_A,GC,s>$ where $p_A\in{}P_A$: \newline
The generation core edits $p_A$ using $s$ towards an instance of class $B$; we denote this operation as $GC(p_A,B,s)$ and its output as $x$.
\item
the \emph{validity} of the transition attempt as a flag in $\{0,1\}$:\newline
A transition attempt is valid (flag=1) if the class predicted by $f_{AB}$ for $x$ is indeed $B$,
otherwise the attempt is not valid (flag=0).
\item
the \emph{plausibility} of the transition attempt as the proximity of $x=GC(p_A,B,s)$ to the closest prototype of $p_B$: \\
We implement this as $sim(x,p^{best}_B)$ where $p^{best}_B=\arg\min_{y\in{P_B}}(sim(x,y))$; as proximity function $sim()$ we use the inverse of the Mahalanobis distance.
\end{itemize}
 Conceptually, the source prototype $p_A$ is the factual we attempt to transport for $A$, and $p^{best}_B$ is the target prototype that serve as anchor after the scan $s$. 

For a transition attempt,
validity determines if the decision boundary of $f_{AB}$ was successfully crossed, while plausibility assesses to what extend that crossing agrees with the true data distribution. A high plausibility value means that the scan $s$ of $GC()$ resulted in an object that is close to one of the target classes prototypes. A low plausibility value means that the counterfactual $x$ strayed away from the data manifold; notice that this is independent of the validity flag.

It is stressed that a transition attempt is not only determined by the prototype chosen from $P_A$ but also from the concrete scan $s$ of the generator core. To derive a proximity between a prototype $p_A$ and the target class $B$ we need to generate multiple transition attempts. To determine the proximity between $A$ and $B$ over all prototypes, we need a further aggregation, which leads to the CP-proximity.


\subsubsection{CP-Proximity.}
We use the same specifications of source class $A$, target class $B$, separating classifier $f_{AB}$ and counterfactual generation core $GC$. We define the \textit{asymmetric proximity} from $A$ to $B$ as the average of the validity flags for all transition attempts towards $B$: we vary folds and retrieval strategies (that determine the scans), and do so for each prototype in $P_A$.

Finally, we define the \textit{CP-proximity} between $A$ and $B$ as the average of all transition attempts, pooling together the attempts from $A$ to $B$ and those from $B$ to $A$,  or equivalently aggregating the two asymmetric proximities from $A$ to $B$ and from $B$ to $A$. This is legitimate, because $f_{AB}$ defines one and only boundary between $A$ and $B$, and the edits cross this boundary independently of the direction. If the prototypes of the one class are more challenging to flip than those of the other class, then the average validity will be lower.
\color{black}




\subsection{The Transition Graph as basis for the evaluation of CF-explainers}
\label{sec:method_cftg}

We build a benchmark for CF-explainers on the basis of the CP-proximity concept. In particular, let $\mathcal{C}$ be a set of $K$ classes. For each pair of classes $A,B$, we train a binary classifier $f_{AB}$ and build the $top-m$ prototypes/medoids over the embeddings space of this classifier as described in Section~\ref{sec:method_prototypes}. This means that we build $\binom{K}{2}$ classifiers. Let $GC$ be the counterfactual generation core we want to evaluate on $\mathcal{C}$.

We span the fully connected \textit{CounterFactual Transition graph} $CFTG(\mathcal{C},GC)$ (or 'Transition Graph' for short), where a node is a class $A\in\mathcal{C}$. An undirected edge $<A,B>$ is a tuple $<P_A,P_B,weight(A,B)>$, where $P_A, P_B$ are the prototypes of $A$ and $B$ in the embeddings space of $f_{AB}$ and $weight(A,B)$ is the CP-proximity between $A,B$ on the basis of the transition attempts of $GC$.

%
We evaluate $GC$ on the CFTG using following quantities:
\begin{itemize}
    \item \emph{CF-rigidity:} this is the 1-complement of CP-proximity for a pair of classes. High CP-rigidity, or equivalently low average validity, means that most of the transition attempts to cross the boundary defined by the binary classifier are invalid. Low CP-rigidity means that most of the transitions are valid, ie the boundary is 'porous' and thus easy to cross.
    \item \emph{Classifier difficulty;} this is the classification error of $f_{AB}$ for classes $A,B$.
    \item \emph{Difficulty/Rigidity interplay:} we quantify this as the Spearman correlation coefficient between CP-proximity and classifier difficulty, computed over the $\binom{K}{2}$ classifiers for all pairs of classes.
\end{itemize}

\color{black}

Figure~\ref{fig:badcf} illustrates the interplay between rigidity and difficulty on the example of a prototype instance. We choose the Fingertip-Touch/Clothes-Peg boundary, because the classifier's difficulty is low, indicating a clear-cut boundary. Among the counterfactual generation cores, replacement methods stay on-manifold but fail to flip the label, while gradient methods succeed only via off-manifold changes.

\section{Experimental Setup}\label{sec:setup}
We evaluate our approach on the SenseGlove DK1.3 recordings of
Section~\ref{sec:materials} (six classes, $34$ subjects, $W{=}150$-sample
windows over the $N{=}20$ node hand graph) under $5$-fold subject-grouped
cross-validation.

\paragraph{Labeling and base classifier:}
We train $15$ classifiers, one per class pair and fold ($75$ models), sharing
the architecture and grouped splits of the six-way model; their mean
validation accuracy is $0.97$. The classifier finds Clothes-Peg against the
small Rubik rotations hardest ($T_1,T_5$ at $0.904$, $T_1,T_6$ at $0.933$)
and the Clothes-Peg against Rubik pairs easiest ($T_1,T_3$, $T_1,T_4 \geq
0.988$). These accuracies constitute the confusion-matrix, which we juxtapose
to the transition graph. 

\paragraph{Counterfactual generation:}
We instantiate the transition operation of
Section~\ref{sec:method_cftg} with the four generators of
Section~\ref{sec:materials}: the replacement-based GACP and GACP-Context, and
the gradient-based NEPO and Residual. For each pair and fold, we select $m{=}3$ prototypes per class ($9$ source-target prototype pairs). We label a transition as 'valid', when the target probability 
exceeds $0.5$.  For the replacement generators we sweep nine retrieval specifications (nearest-$k$ for $k\in\{1,3,5,10\}$, percentiles $25/50/75$, furthest-$5$, random-$5$) that trace the CF-proximity axis, while the gradient generators run once per prototype pair. This yields $900$ attempts per class pair over the five folds, and $13{,}500$ across the graph. All counterfactuals are computed on held-out validation windows the relevant classifier already predicts correctly. To enforce kinematic consistency, we modified only the $N{=}20$ raw joint-angle channels, recalculating velocity and acceleration accordingly. 

\paragraph{CFTG Construction:}
We build the CFTG on top of the Physio-Digital Temporal Graph (PDTG) classifier $f$ of Section~\ref{sec:materials}~\cite{ZaidiEtAl:CBMS2026}, which distinguishes the six hand movements, and read it through the four generation cores of the same section. Because a counterfactual question is contrastive (it moves from one class to another class), we instantiate the explanation task on the $\binom{K}{2}=15$ binary specialisations $f_{AB}$ of $f$, one per class pair, each sharing its architecture and training protocol. This removes the multi-target ambiguity of a native multi-class counterfactual (toward which of the other classes should the edit head?) and gives every edge of the CFTG a single-boundary meaning.

\paragraph{Note:}
The workflow described here involves several design choices, including number of prototypes, retrieval settings, and distance metrics. We have built upon insights from \cite{ZaidiEtAl:CBMS2026}. A systematic sensitivity analysis of these choices is outside the present scope and is planned as future work.

    \section{Results}\label{sec:results}

We build the CFTG over the six classes from the $13{,}500$ transition
attempts of Section~\ref{sec:setup} and read its structure against the
classifier's accuracy. Table~\ref{tab:edges} lists, per class pair, the
binary accuracy, the pooled edge weight $w$ (mean validity), the
plausibility of the successful flips, and the per-generator validity, sorted
from the most rigid boundary to the most porous one.

\begin{table}[t]
\centering
\caption{Quantities on the edges of the transition graph, one row per
class pair, sorted by increasing CP-proximity $w$ . ``Acc'' is accuracy (1-classifier difficulty); ``Plaus.'' is the mean plausibility (lower is more realistic); the last four columns give per-generator validity.
}
\label{tab:edges}
\small
\setlength{\tabcolsep}{5pt}
\begin{tabular}{@{}lccccccc@{}}
\toprule
Pair & Acc & $w$ & Plaus. & GACP & GACP-Ctx & NEPO & Residual \\
\midrule
$T_1,T_2$ & 0.985 & \textbf{0.08} & 10.9 & 0.00 & 0.02 & 0.80 & 0.53 \\
$T_1,T_3$ & 0.988 & \textbf{0.08} & 15.9 & 0.05 & 0.01 & 0.33 & 0.80 \\
$T_1,T_4$ & 0.989 & \textbf{0.11} & 23.5 & 0.04 & 0.01 & 0.87 & 0.93 \\
$T_2,T_3$ & 0.987 & \textbf{0.12} & 8.5  & 0.04 & 0.02 & 1.00 & 0.80 \\
$T_4,T_5$ & 0.972 & 0.26 & 7.2  & 0.20 & 0.21 & 0.93 & 0.47 \\
$T_1,T_6$ & 0.933 & 0.26 & 7.0  & 0.17 & 0.22 & 0.73 & 1.00 \\
$T_3,T_4$ & 0.983 & 0.30 & 6.3  & 0.17 & 0.32 & 0.80 & 0.73 \\
$T_3,T_5$ & 0.974 & 0.34 & 4.2  & 0.26 & 0.29 & 1.00 & 0.87 \\
$T_2,T_6$ & 0.968 & 0.34 & 2.9  & 0.12 & 0.46 & 0.87 & 0.73 \\
$T_5,T_6$ & 0.941 & 0.35 & 6.4  & 0.02 & 0.54 & 1.00 & 1.00 \\
$T_4,T_6$ & 0.964 & 0.38 & 4.7  & 0.16 & 0.48 & 0.93 & 0.93 \\
$T_1,T_5$ & 0.904 & 0.42 & 8.5  & 0.36 & 0.38 & 0.73 & 0.93 \\
$T_3,T_6$ & 0.971 & 0.49 & 2.9  & 0.33 & 0.57 & 0.80 & 0.93 \\
$T_2,T_4$ & 0.989 & 0.49 & 3.5  & 0.49 & 0.39 & 1.00 & 0.93 \\
$T_2,T_5$ & 0.980 & \textbf{0.56} & 3.2  & 0.46 & 0.60 & 0.80 & 0.80 \\
\midrule
mean & 0.969 & 0.31 & 7.6 & 0.19 & 0.30 & 0.84 & 0.83 \\
\bottomrule
\end{tabular}
\end{table}

\subsection{Transition graph versus confusion matrix}
\label{sec:res_topology}

\begin{figure}[!htbp]
\centering
\includegraphics[width=1\linewidth]{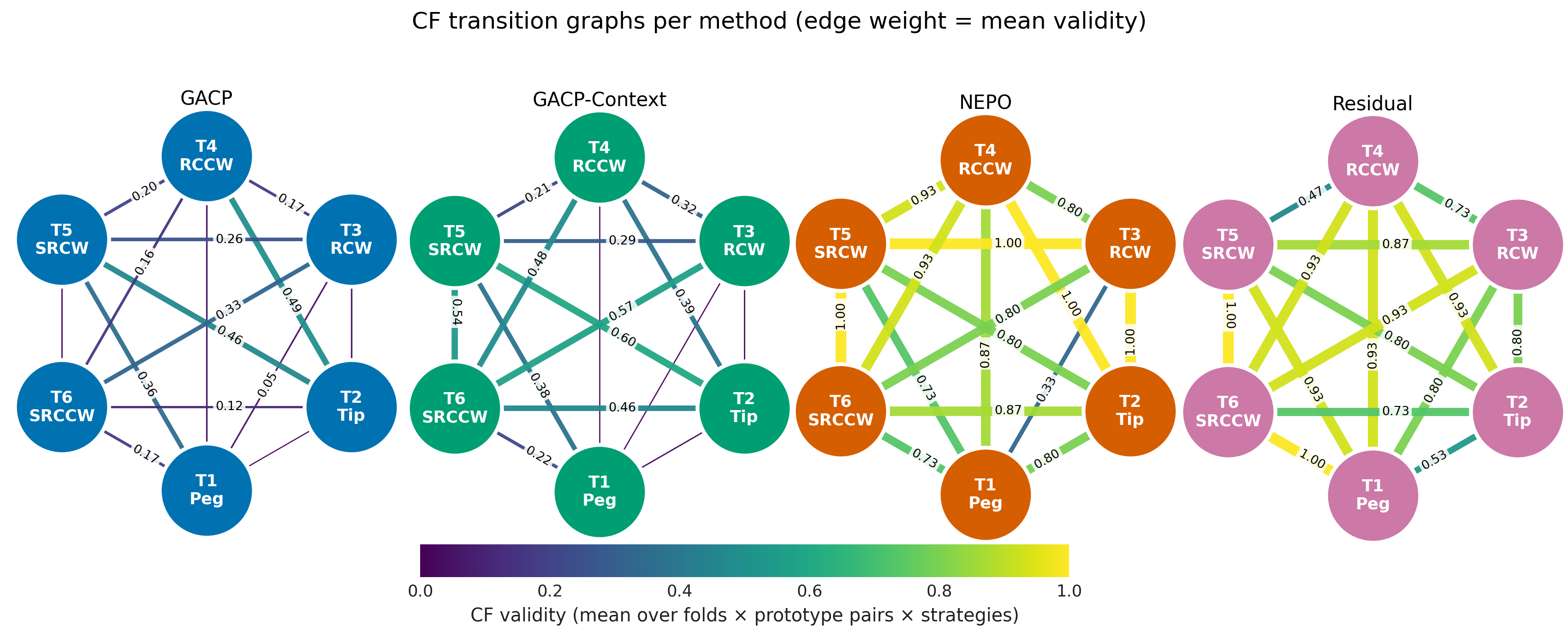}
\caption{The counterfactual transition graph computed separately for each
generator. Nodes are classes; each edge weight (colour and thickness) is the
mean validity of that boundary. The replacement generators (GACP,
GACP-Context; top) leave every $T_1$ (Clothes-Peg) boundary thin and dark and
recover a sparse, structured topology, whereas the gradient generators (NEPO,
Residual; bottom) flood almost every boundary, reaching all classes by
leaving the data manifold. Pooling the four panels gives the edge weights $w$
analysed in Table~\ref{tab:edges}.}
\label{fig:cftg}
\end{figure}

Figure~\ref{fig:cftg} shows the transition graph for each generator. The four panels share the six-class layout, and edge colour and thickness encode the rigidity of that boundary. Pooling the generators gives the edge values of Table~\ref{tab:edges}.
The most porous boundaries are Fingertip-Touch against the small Rubik rotation ($T_2,T_5$; $w{=}0.56$) and against Rubik counter-clockwise ($T_2,T_4$; $0.49$), and Rubik clockwise against Small-Rubik counter-clockwise ($T_3,T_6$; $0.49$). The most rigid boundaries
all involve Clothes-Peg ($T_1$): a central $T_1$ window is almost never moved to Fingertip-Touch, Rubik clockwise or Rubik counter-clockwise by an in-distribution edit ($w{=}0.08$, $0.08$, $0.11$). 
Results also depend on the generator: replacement graphs are sparse, while gradient-based graphs are dense (see Section~\ref{sec:res_methods}).

\begin{figure}[htb]
\centering
\includegraphics[width=0.72\linewidth]{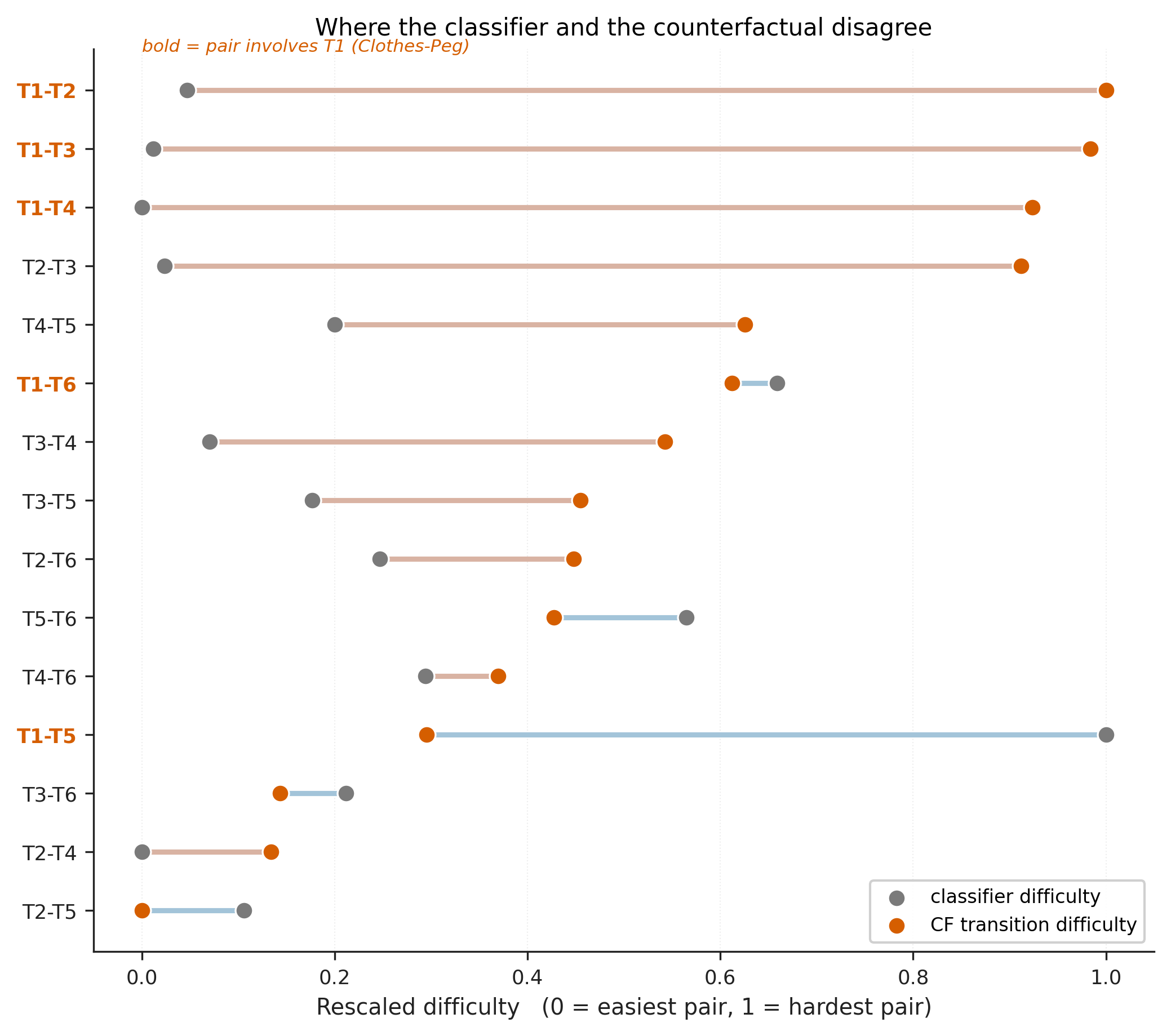}
\caption{Classifier difficulty (grey) vs.\ CF rigidity (orange) per class pair
after min-max rescaling of each quantity independently. For both quantities, values closer to zero are better - easier separation, resp. easier class flipping.
The long top connectors ($T_1,T_2$; $T_1,T_3$; $T_1,T_4$) are boundaries the classifier
separates almost perfectly yet a realistic edit almost never crosses; $T_1,T_5$
inverts this. 
}
\label{fig:dumbbell}
\end{figure}

On Figure \ref{fig:dumbbell} we juxtapose classifier difficulty (gray dots) to rigidity (orange dots). These quantities have different native scales, so we rescaled them (min-max) \emph{independently} to $[0,1]$: a dot's position is the pair's \emph{rank} between the easiest ($0$) and hardest ($1$) boundary on that axis, 
so the panel compares orderings, not magnitudes. Pairs are sorted by CF rigidity (hardest at top); bold labels involve Clothes-Peg ($T_1$), and connector length is the ranking disagreement. 


In the Figure, we see that while classifier difficulty is low for all pairs, i.e. all bluish dots are closer to zero than to one, the CF-difficulty goes in the opposite direction. In particular, the pairs separated \emph{most} confidently by the classifier are the ones that are \emph{least} often crossed by in-distribution CFs. This can be seen at the pairs T1,T4 and T1,T3: they have very low classifier difficulty, while the CF-difficulty is maximal. This is reversed for the pair T1,T5: difficult class separation (bluish dot close to one) but low CF-difficulty (orange dot is at a position between 0.2 and 0.3). We intentionally point to pairs that involve class T1 (Clothes-Peg): the T1 gesture is very different from the other ones, and this is more challenging for the explainers than for the classifier.

Finally, we computed the Spearman correlation coefficient between classifier difficulty and rigidity
for the 15 pairs of classes, and we found a 
weak negative rank association (Spearman $\rho{=}-0.37$, $p{=}0.18$). Since the number of pairs is small, and since we did not correct for multiple testing (each gesture appears in more than one pairs), we cannot reach statistical significance. Nonetheless, the trend is clear and disquieting: if we concentrate on the extremes, i.e. the longest lines in Figure \ref{fig:dumbbell}, the discrepancy in difficulty suggests that the easier the two classes are to separate, the harder is to get the valid counterfactual for them!


\subsection{A singleton cluster and non-taxonomic clusters}
\label{sec:res_clusters}
We build a cluster dendrogramm of the gesture classes on rigidity and found that the first split
isolates Clothes-Peg ($T_1$) as a \emph{singleton}: no other class is
CF-near to it, although the pairs $T_1,T_3$ and $T_1,T_4$ are the easiest to separate.

The next level of the dendrogramm is surprising: the  six tasks were organised by humans along object size and rotation direction ($T_3,T_4$ large-cube clockwise/counter-clockwise, $T_5,T_6$ small-cube, $T_1,T_2$ peg and fingertip), so one might expect same-object or
same-direction pairs to be the nearest. Instead, the clusterer places
$T_3$ (\text{Rubik CW}), $T_6$ (\text{Small-Rubik CCW}) in one cluster and
$T_2$ (\text{Tip}), $T_4$ (\text{Rubik CCW}) and $T_5$ (\text{Small-Rubik CW}) in the other,
mixing sizes and directions. The same-object opposite-direction pairs are
only moderately porous ($w{=}0.30$ for $T_3,T_4$, $0.35$ for $T_5,T_6$).


\subsection{On-manifold and off-manifold transitions}
\label{sec:res_methods}
To test the generators themselves, we build
per-generator graphs of Fig.~\ref{fig:cftg} and
the per-generator columns of Table~\ref{tab:edges}. The two
gradient GCs flip almost everything (mean validity $0.84$ for NEPO,
$0.83$ for Residual) and cross even the rigid Clothes-Peg boundaries, whereas
the two replacement GCs flip far less ($0.19$ for GACP, $0.30$ for
GACP-Context) and collapse to near zero on exactly those boundaries. The
plausibility column of Table~\ref{tab:edges} explains why: on the rigid pairs the successful
flips are highly implausible (at-flip Mahalanobis $23.5$ for $T_1,T_4$ and
$15.9$ for $T_1,T_3$, versus $2.9$ to $3.5$ on the non-rigid Fingertip-Touch
pairs), and these large distances come almost entirely from the gradient
methods (on $T_1,T_3$, NEPO sits near $50$, GACP near $6$). The gradient
generators reach almost any class, but on rigid boundaries only by stepping
off the data manifold; the replacement generators stay on it and therefore
fail where no in-distribution path exists. The pooled edge weight, dominated
by the in-distribution attempts, is the on-manifold reachability we want, and
the gradient generators supply the off-manifold contrast. This answers RQ1
and RQ2: plausibility, read as distance off the target manifold, captures a
counterfactual's implausibility, and it is large exactly on the rigid
boundaries, that is, between the most dissimilar class prototypes.

The same view separates conclusions about the classifier from artefacts of
the explainer. The generators disagree most on $T_2,T_3$, $T_1,T_4$ and
$T_5,T_6$ (across-generator standard deviation $\approx 0.5$), where the
gradient methods report ``trivial'' and the replacement methods
``impossible''; these are precisely the boundaries a flip can cross only off
the manifold. Pairs on which the generators agree form a method-robust
skeleton that can be read with confidence, whereas high-disagreement pairs
should be reported as generator-sensitive rather than collapsed into one
number. The Fingertip-Touch to Clothes-Peg transition of
Fig.~\ref{fig:badcf} is one such case: in isolation it is one more
counterfactual, but in the graph it is one dark edge into the $T_1$
attractor.

\section{Discussion and Conclusion}\label{sec:disc}

We introduced the counterfactual transition graph as instrument for the evaluation of counterfactual explainers. Its nodes are classes of the dataset we want to evaluate on, each edge aggregate the proportion of valid transition attempts between the two nodes/classes, where the attempts are performed by the counterfactual generation core we want to evaluate. We performed a proof-of-concept evaluation with the SenseGlove dataset of hand gestures, some pairs of which are very easy to separate. Our results revealed that high classifier performance does not imply high quality of the counterfactuals.

\paragraph{Limitations and future work:} We tested our approach on multiple CF generator cores but only on one dataset. In benchmarks, the choice of dataset is pivotal; our dataset has some easily separable class pairs, and valid vs invalid transition attempts can be visualized clearly. We intend to collect further datasets with these properties.
A further limitation concerns the hyperparameter values, especially the number of prototypes per class. We fixed this number to $m=3$ for all classes. The effect of less (or slightly more) prototypes per class, and the influence of the dataset size, remain open for future work.

\color{black}


\paragraph{Outlook:}
The natural next steps are native multi-class counterfactuals in place of pairwise specialisations and a second time-series benchmark.
A further direction, consistent with our view, would be to exploit the pairwise transition graph to construct an ensemble counterfactual explainer, treating each class-to-class transition
as a separate decision problem.

\begin{credits}
\subsubsection{\ackname} 
The work of the first author was financed by the graduate school \textit{TACTIC: “Towards Co-Evolution in Human-Technology Interfaces”}. TACTIC is co-funded by the Federal state of Saxony-Anhalt and the EU (ESF+), Funding Nr. ZS/2023/12/182067. We express our deepest gratitude to \textit{Dr. Christoph Reichert, Dr. Elena Azanon, Lisa Klemm} (Leibniz-Institut für Neurobiologie) for providing the data and constructive feedback on this research. Generative AI (Claude) was used to assist in drafting portions of this manuscript; the authors reviewed, edited, and take full responsibility for the final content.

The work of Szymon Bobek has been supported by a grant from the Priority Research Area (DigiWorld) under the Strategic Programme Excellence Initiative at Jagiellonian University.
This paper is part of a project that has received funding from the European Union's Horizon Europe Research and Innovation Programme, under Grant Agreement number 101120406. The paper reflects only the authors' view and the EC is not responsible for any use that may be made of the information it contains.

\subsubsection{\discintname}
The authors have no competing interests to declare that are relevant to the content of this article.
\end{credits}

\bibliographystyle{plain}
\bibliography{bibKinematics}

\newpage

\appendix

\end{document}